\documentclass[letterpaper,10pt,conference]{ieeeconf}
\IEEEoverridecommandlockouts %
\usepackage{amsmath,amssymb,bm}
\usepackage{graphicx}
\usepackage{booktabs,multirow,makecell,tabularx}
\usepackage[dvipsnames]{xcolor}
\usepackage{algorithm}
\usepackage{algpseudocode}
\usepackage{cite}
\usepackage{hyperref}
\hypersetup{hidelinks,
  pdftitle={Learning Wheel Motion Uncertainty for Degeneracy-Aware LiDAR-Inertial Odometry on Quadrupedal Wheel-Legged Robots},
  pdfauthor={Hairuo Wei, Yunfan Ren, Renda Yang, Qiangsheng Huang, Zhongyu Li}}

\makeatletter
\long\def\@makecaption#1#2{%
  \ifx\@captype\@IEEEtablestring
    {\footnotesize\centering #1\par
      {\scshape #2\par}}%
    \vskip 3pt
  \else
    \vskip\abovecaptionskip
    \parbox[t]{\hsize}{\footnotesize\raggedright
      \noindent #1.\enspace #2\par}%
  \fi}
\makeatother
\AtBeginDocument{%
  \setlength{\abovedisplayskip}{6pt plus 2pt minus 2pt}%
  \setlength{\belowdisplayskip}{6pt plus 2pt minus 2pt}%
  \setlength{\abovedisplayshortskip}{4pt plus 2pt minus 2pt}%
  \setlength{\belowdisplayshortskip}{4pt plus 2pt minus 2pt}%
}
\begin{document}
\raggedbottom

  \title{\LARGE\bf Learning Wheel Motion Uncertainty for Degeneracy-Aware
    LiDAR-Inertial Odometry on Quadrupedal Wheel-Legged Robots}

\author{Hairuo Wei$^{1}$, Yunfan Ren$^{2}$, Renda Yang$^{3}$,
Qiangsheng Huang$^{3}$, and Zhongyu Li$^{1,*}$%
\thanks{$^{1}$The Chinese University of Hong Kong, Hong Kong SAR, China.}%
\thanks{$^{2}$Tongji University, Shanghai, China.}%
\thanks{$^{3}$SEER Robotics.}%
\thanks{\parbox[t]{\dimexpr\linewidth-\parindent\relax}{%
\raggedright $^{*}$Corresponding authors:\protect\newline
hairuo@connect.hku.hk; zhongyu\_li@berkeley.edu.}}%
}

\maketitle
\thispagestyle{empty}
\pagestyle{empty}
\begin{abstract}
    Wheel kinematics can support odometry for quadrupedal wheel-legged robots when LiDAR geometry provides weak motion constraints.
    However, slip and contact loss make rolling predictions unreliable, and contact state alone cannot determine their reliability across motion directions.
    We propose a tightly coupled LiDAR--inertial--kinematic odometry framework that retains analytical rolling predictions and learns per-wheel, direction-dependent process covariances from proprioceptive history.
    The estimator maintains moving wheel-center states, allowing the learned uncertainty to regulate the influence of rolling predictions through forward-kinematic corrections.
    Degeneracy-aware projection (DAP) complements this mechanism by suppressing LiDAR constraints along geometrically weak directions.
    The learned covariance network (CovNet) is trained through a differentiable filter using simulated velocity supervision and deployed on the real robot without fine-tuning.
    On the six room sequences with the highest degeneracy ratios in our indoor dataset, the method reduces mean per-sequence position root mean square error (RMSE) by 66.2\% relative to FAST-LIO2 and by 31.0\% relative to a fixed-covariance variant.
    In a severely degenerate corridor run, it achieves an endpoint position error of 0.198~m, compared with 7.531~m for FAST-LIO2.
    These results demonstrate robust odometry under challenging geometric and wheel-motion conditions.
    We plan to publicly release our code and multimodal dataset.
    Project website: {\urlstyle{same}\url{https://weihairuo.github.io/wheel-motion-uncertainty/}}.
\end{abstract}

\section{Introduction}
\label{sec:intro}

\begin{figure}[!t]
    \centering
    \includegraphics[width=\columnwidth]{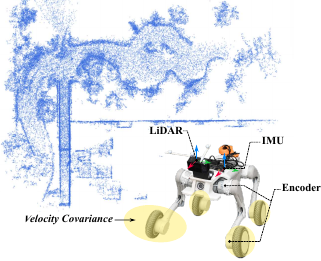}
    \footnotesize\frenchspacing
    \spaceskip=\fontdimen2
    \caption{Quadrupedal wheel-legged state estimation using proposed method,
    which fuses LiDAR, inertial measurement unit (IMU), and encoder
    measurements together with learned wheel-motion uncertainty.
    Yellow ellipsoids represent covariance.}
    \label{fig:cover}
\end{figure}

Quadrupedal wheel-legged robots (Fig.~\ref{fig:cover}) combine efficient rolling with the ability to traverse uneven terrain \cite{wen2025wheellegged,liu2025terrain}.
Reliable state estimation is essential for their control, navigation, and mapping.
LiDAR--inertial odometry (LIO) provides accurate motion estimates when the environment contains sufficient geometric features \cite{xu2021fast,xu2022fastlio2}.
However, corridors, tunnels, and open areas may provide weak geometric constraints along certain motion directions.
This can make LiDAR registration ill-conditioned and sensitive to noise \cite{hu2026dcreg}.
Joint and wheel encoders provide an additional source of motion information that can support state estimation in these conditions.

Using wheel kinematics to support LIO on quadrupedal wheel-legged robots presents two related challenges.
First, the reliability of each wheel's rolling prediction varies with its ground interaction and across motion directions.
Conventional wheel odometry commonly assumes a rigid chassis \cite{yuan2023liwo}, while contact-aided legged estimators often assume stationary stance feet \cite{bloesch2013state,hartley2020contact}.
For articulated wheel-legs, forward kinematics (FK) describes each wheel center's position relative to the body, while a rolling model predicts its motion in the \textcolor{black}{global frame}.
Slip, contact loss, and unloaded wheel spin can reduce the reliability of this prediction without invalidating the relative position obtained from FK.
Contact state alone is insufficient to determine reliability: a supported wheel may still slip, with different errors across motion directions.
Second, LiDAR updates must account for direction-dependent geometric weakness while retaining useful measurement information.
Reliable wheel-motion constraints can support estimation along geometrically weak directions, while informative LiDAR measurements can limit drift when rolling predictions become unreliable.
Exploiting this complementarity requires accounting for both wheel-motion uncertainty and the directional weakness of LiDAR constraints.

Existing methods incorporate wheel constraints and adaptive weighting \cite{yuan2023liwo,guadagnino2025kinematic}, estimate leg-odometry bias \cite{wisth2022vilens}, or introduce wheel-legged kinematic and terrain models \cite{wen2025wheellegged,liu2025terrain}.
Learning-based approaches estimate contact states, correct kinematic measurements, or learn motion models online \cite{lin2022legged,youm2024neural,okawara2025wheel,okawara2025tightly}.
Differentiable filtering also enables uncertainty models to be learned from state-estimation objectives \cite{haarnoja2016backprop,kloss2021differentiable,baumgartner2026coco}.
For a wheel center, nominal motion persists even during ideal rolling, so uncertainty must describe deviations from that motion.
The resulting wheel-motion constraints must be fused with LiDAR information whose reliability varies across pose directions.

\textcolor{black}{We propose a tightly coupled LiDAR--inertial--kinematic odometry framework for quadrupedal wheel-legged robots.
The framework combines analytical rolling predictions with learned wheel-motion uncertainty and degeneracy-aware LiDAR updates.
The main contributions of this work are:}

\begin{itemize}
    \item \textcolor{black}{A tightly coupled LiDAR--inertial--kinematic estimator with persistent wheel-center states, reducing mean per-sequence position root mean square error (RMSE) by 66.2\% relative to FAST-LIO2 and by 40.4\% relative to the model-based NCDM baseline on the six room sequences with the highest degeneracy ratios.}
    \item \textcolor{black}{A learning-based approach that estimates per-wheel, direction-dependent rolling-motion uncertainty from proprioceptive history to adaptively weight analytical rolling predictions.}
    \item \textcolor{black}{A degeneracy-aware projection (DAP) for LiDAR updates that suppresses direct information along geometrically weak directions while preserving informative constraints.}
    \item \textcolor{black}{Extensive real-robot experiments under geometric degeneracy, wheel slip, and external disturbances, supported by component ablations. We will publicly release a dataset collected with our quadrupedal wheel-legged robot.}
\end{itemize}

\section{Related Work}
\label{sec:related}

\textcolor{black}{For robust state estimation of quadrupedal wheel-legged robots, we review how prior work incorporates kinematic constraints, models their uncertainty, and handles LiDAR degeneracy.}

\subsection{Kinematics-Aided State Estimation}

Robot kinematics provides motion information when environmental features are insufficient for reliable LiDAR registration.
For wheeled robots, LIWO incorporates encoder-derived velocity observations into LiDAR--inertial estimation \cite{yuan2023liwo}, while Kinematic-ICP constrains registration using platform kinematics and adapts the influence of wheel odometry \cite{guadagnino2025kinematic}.
Online neural kinematic models further adapt wheel-motion predictions to changing terrain conditions \cite{okawara2025wheel}.
For articulated robots, VILENS combines leg odometry with visual, inertial, and LiDAR measurements and estimates a velocity bias to reduce kinematic errors \cite{wisth2022vilens}, while a learned leg-kinematics model uses tactile inputs to adapt proprioceptive constraints \cite{okawara2025tightly}.

Wheel-legged estimation must additionally account for rolling motion.
Existing systems introduce kinematic corrections \cite{wen2025wheellegged} and terrain constraints \cite{liu2025terrain}, while contact-anchored odometry propagates contact positions using wheel rolling displacement \cite{sun2026contact}.
Zhou et al. combine probabilistic slip modeling, offset wheel-contact geometry, and generalized-momentum-based contact detection in a proprioceptive multimode estimator \cite{zhou2025multimode}.
Our formulation maintains all four moving wheel centers as persistent states and assigns learned process covariances to their analytical rolling predictions.
FK measurements couple these states to the body within the same estimator that incorporates LiDAR information, allowing wheel-motion uncertainty to influence the fused state estimate.

\subsection{Contact Modeling and Learned Uncertainty}

The usefulness of a kinematic constraint depends on the underlying ground interaction.
Contact-aided extended Kalman filters (EKFs) use stance feet as approximately stationary references and relate them to the body through FK measurements \cite{bloesch2013state}.
This formulation has been extended to invariant filtering \cite{hartley2020contact} and factor graphs with preintegrated contact motion or contact forces \cite{hartley2018legged,fourmy2021contact}.
Camurri et al. estimate contact probabilities and detect impacts from internal force sensing, using this information to weight individual legs in velocity estimation \cite{camurri2017probabilistic}.
Learning-based contact predictors determine when a constraint should be used \cite{lin2022legged}, while neural measurement models correct the motion information supplied to the estimator \cite{youm2024neural}.
For rolling wheels, contact alone does not establish the reliability of the predicted motion: a supported wheel may still slip, with errors that differ across directions.

Differentiable filtering provides a way to learn uncertainty models through their effect on state estimation.
Backprop KF jointly trains an observation network and its predicted observation covariance through Kalman filtering using state-estimation losses \cite{haarnoja2016backprop}.
Kloss et al. study end-to-end training of differentiable filters, including the learning of process and observation noise models whose covariances vary with the inputs \cite{kloss2021differentiable}.
For legged robots, CoCo-InEKF learns directional contact-velocity covariances through a differentiable filter, without contact labels \cite{baumgartner2026coco}.
\textcolor{black}{CoCo-InEKF learns directional confidence in a zero-mean contact-velocity model.
For wheel-legged robots, however, wheel centers move even during ideal rolling.
Our formulation explicitly models this nominal motion and learns its uncertainty, allowing reliable rolling constraints to support state estimation, with real-world ablation experiments showing a 31.0\% reduction in mean per-sequence position RMSE (Table~\ref{tab:exp_mocap}).}

\subsection{Degeneracy-Aware LiDAR Estimation}

Reliable wheel constraints can support estimation when LiDAR geometry is weak, but the LiDAR update must also account for that weakness.
Earlier work identifies poorly constrained directions and restricts the corresponding estimation updates \cite{zhang2016degeneracy}.
X-ICP evaluates how scan-to-map correspondences constrain the principal optimization directions and incorporates the resulting localizability categories into constrained registration \cite{tuna2024xicp}.
Hatleskog and Alexis account for noise in points and surface normals when estimating the probability of degeneracy, then smoothly attenuate updates in unreliable directions \cite{hatleskog2024probabilistic}.
Tuna et al. further compare truncated singular value decomposition, inequality constraints, and regularization in field experiments, examining how these choices affect ill-conditioned registration \cite{tuna2025informed}.
DCReg uses Schur complements to separate rotational and translational constraints when characterizing ill-conditioning \cite{hu2026dcreg}.
Our estimator uses DAP to screen the full pose-information matrix and project the LiDAR Jacobian onto retained directions.
DAP controls direct LiDAR information from weak geometry, while learned wheel-process covariances control uncertainty in the rolling predictions.
These components act through a shared state and covariance, coupling the retained LiDAR information to the IMU--wheel-center prior.

\section{Methodology}
\label{sec:method}

\subsection{System Overview}
\label{sec:method_overview}

\begin{figure}[!t]
    \centering
    \includegraphics[width=\columnwidth]{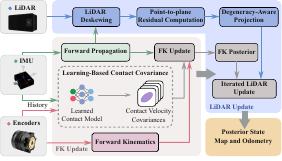}
    \caption{Overview of the proposed system.
        Given IMU, encoder, and LiDAR measurements,
        a learned covariance network (CovNet) predicts wheel-motion process covariances
        from proprioceptive history, while
        degeneracy-aware projection (DAP) suppresses
        direct LiDAR information along
        geometrically weak directions
        during the LiDAR update.
        The system then fuses these measurements
        to estimate the robot state
        and build a point cloud map.}
    \label{fig:system_overview}
\end{figure}

To estimate robot motion and build a point cloud map, our system
fuses IMU, encoder, and LiDAR measurements in an error-state
iterated Kalman filter (ESIKF), as illustrated in
Fig.~\ref{fig:system_overview}.
Rolling wheel-center predictions and encoder-based \textcolor{black}{forward kinematics (FK)} observations
provide motion constraints when LiDAR geometry is weak
(Sec.~\ref{sec:wheel_process}).
Because terrain and directional slip change the reliability of
these predictions, a learned covariance network (CovNet) uses
proprioceptive history to predict
time-varying wheel-process covariances (Sec.~\ref{sec:learned_uncertainty}).
The filter performs FK and LiDAR updates sequentially, using the
FK posterior as the LiDAR prior.
LiDAR constrains drift when rolling is unreliable, while
degeneracy-aware projection (DAP)
suppresses direct LiDAR information along geometrically weak
directions (Sec.~\ref{sec:geometry_lidar}).
\par

\subsection{\textcolor{black}{Forward Kinematics} Update}
\label{sec:wheel_process}
\label{sec:joint_estimation}

\textcolor{black}{This module fuses IMU and encoder 
measurements within the ESIKF.
We first define the state representation, 
then present the process model and the rolling kinematics 
used for prediction.
Finally, we formulate the measurement update using encoder
observations to correct the predicted state and covariance.}

\subsubsection{State Representation}
Since the IMU is rigidly attached to the robot body, we use the IMU frame
$I$ as the body reference frame throughout this section.
We take the initial IMU frame as the fixed global frame $G$.
We retain all four wheel centers in the filter state to incorporate
rolling motion and FK observations. With fixed sensor extrinsics,
the state is
\begin{equation}
    \mathbf x = \bigl({}^{G}\mathbf R_I,\!{}^{G}\mathbf p_I,\!{}^{G}\mathbf v_I,
    \!{}^{I}\mathbf b_g,\!{}^{I}\mathbf b_a,\!{}^{G}\mathbf g,
    \!{}^{G}\mathbf p_{c_1},\!\ldots,\!{}^{G}\mathbf p_{c_4}\bigr),
    \label{eq:online_state}
\end{equation}
where ${}^{G}\mathbf R_I$, ${}^{G}\mathbf p_I$, and ${}^{G}\mathbf v_I$
respectively denote the IMU attitude, position, and velocity in the global
frame. The states ${}^{I}\mathbf b_g$ and ${}^{I}\mathbf b_a$ denote
the gyroscope and accelerometer biases in the IMU frame, respectively.
${}^{G}\mathbf g$ denotes the gravity vector in the global frame, and
${}^{G}\mathbf p_{c_i}$ denotes the position of wheel center $i$ in the
global frame, with $i\in\{1,\ldots,4\}$. The local estimation error is
denoted by $\delta\mathbf x\in\mathbb R^{30}$, with covariance
$\mathbf P\in\mathbb R^{30\times30}$.

\subsubsection{IMU and Wheel-Center Process Model}
We use standard IMU dynamics~\cite{xu2022fastlio2} and augment them
with rolling wheel-center dynamics:
{\allowdisplaybreaks[1]
\begin{align}
    {}^{G}\dot{\mathbf R}_I
                            & = {}^{G}\mathbf R_I
    \left[{}^{I}\boldsymbol\omega_m-{}^{I}\mathbf b_g
    -{}^{I}\mathbf n_g\right]_\times,
    \label{eq:joint_process_model}                                                                     \\
    {}^{G}\dot{\mathbf p}_I & = {}^{G}\mathbf v_I,
    \label{eq:position_process}                                                                        \\
    {}^{G}\dot{\mathbf v}_I
                            & = {}^{G}\mathbf R_I
    \left({}^{I}\mathbf a_m-{}^{I}\mathbf b_a-{}^{I}\mathbf n_a\right)
    +{}^{G}\mathbf g,
    \label{eq:velocity_process}                                                                        \\
    {}^{I}\dot{\mathbf b}_g & = {}^{I}\mathbf n_{b_g},
    \label{eq:gyro_bias_process}                                                                       \\
    {}^{I}\dot{\mathbf b}_a & = {}^{I}\mathbf n_{b_a},
    \label{eq:accel_bias_process}                                                                      \\
    {}^{G}\dot{\mathbf g}   & = \mathbf 0,
    \label{eq:gravity_process}                                                                         \\
    {}^{G}\dot{\mathbf p}_{c_i}
                            & = {}^{G}\mathbf R_I\left({}^{I}\mathbf u_i+{}^{I}\mathbf n_{w_i}\right),
    \quad i=1,\ldots,4.
    \label{eq:wheel_center_process_overview}
\end{align}
}
Here, ${}^{I}\boldsymbol\omega_m$ and ${}^{I}\mathbf a_m$ are the gyroscope
and scale-calibrated accelerometer measurements. The terms ${}^{I}\mathbf n_g$
and ${}^{I}\mathbf n_a$ denote their measurement noise, while
${}^{I}\mathbf n_{b_g}$ and ${}^{I}\mathbf n_{b_a}$ denote bias process noise.
Gravity is modeled as constant during propagation. For each wheel,
${}^{I}\mathbf u_i$ is the nominal wheel-center velocity computed from the
rolling model below. The zero-mean white process noise
${}^{I}\mathbf n_{w_i}$ has spectral density ${}^{I}\boldsymbol\Sigma_i$.

\subsubsection{Rolling Kinematics}
\begin{figure}[!t]
    \centering
    \includegraphics[width=0.90\columnwidth]{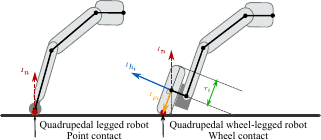}
    \caption{Point contact (left) and wheel contact (right). The effective
        radius $r_i$, wheel axis ${}^{I}\mathbf h_i$, and estimated normal
        ${}^{I}\mathbf n$ define the assumed contact geometry.
        \textcolor{black}{The vector ${}^{I}\boldsymbol\rho_i$ points from the
            wheel center to the assumed contact point.}}
    \label{fig:wheel_motion}
\end{figure}

We use a no-slip rolling constraint to obtain the nominal wheel-center
velocity~\cite{bjelonic2019keep}. As illustrated in Fig.~\ref{fig:wheel_motion},
${}^{I}\boldsymbol\rho_i$ points from the wheel center to the assumed contact
point. We compute this vector as
\begin{equation}
    {}^{I}\boldsymbol\rho_i
    = -r_i\frac{
    \left(\mathbf I_3-{}^{I}\mathbf h_i({}^{I}\mathbf h_i)^\top\right){}^{I}\mathbf n
    }{
    \left\|\left(\mathbf I_3-{}^{I}\mathbf h_i({}^{I}\mathbf h_i)^\top\right){}^{I}\mathbf n\right\|
    },
    \label{eq:wheel_contact_vector}
\end{equation}
where $r_i$ is the effective wheel radius, ${}^{I}\mathbf h_i$ is the
wheel-axis unit vector, $\mathbf I_3$ is the $3\times3$ identity matrix,
and ${}^{I}\mathbf n$ is the upward terrain normal.
We compute ${}^{I}\mathbf h_i$ from the robot kinematic model and measured
leg joint positions $\mathbf q^{\mathrm{leg}}\in\mathbb R^{12}$,
which exclude the wheel joints, using the fixed body-to-IMU extrinsic rotation, and
estimate ${}^{I}\mathbf n$ by fitting a plane to the four FK wheel centers.
Equation~\eqref{eq:wheel_contact_vector} projects the negative terrain
normal onto the wheel plane perpendicular to ${}^{I}\mathbf h_i$, then
normalizes the projection and scales it to $r_i$ to obtain the contact
vector.

For leg $i$, the effective rolling rate is
\begin{equation}
    \omega_i^{\mathrm{roll}}=\sum_{j\in\mathcal J_i}
    \bigl(({}^{I}\mathbf h_i)^\top{}^{I}\mathbf h_{i,j}(\mathbf q^{\mathrm{leg}})\bigr)\dot q_{i,j},
    \label{eq:rolling_rate}
\end{equation}
where $\mathcal J_i$ contains the joints along leg $i$, including the wheel
joint, and ${}^{I}\mathbf h_{i,j}$ denotes the unit axis of joint $j$
expressed in the IMU frame. The scalar $\dot q_{i,j}$ is the measured
rate of this joint, corresponding to an entry of the full joint-rate
vector $\dot{\mathbf q}\in\mathbb R^{16}$.
Since leg articulation also contributes to
wheel rotation relative to the body, we project each joint's angular
velocity onto the wheel axis ${}^{I}\mathbf h_i$ and sum these
contributions to obtain the effective rolling rate.
Using this axial approximation, the nominal wheel-center velocity is
\begin{equation}
    {}^{I}\mathbf u_i
    = {}^{I}\boldsymbol\rho_i\times
    \bigl({}^{I}\boldsymbol\omega_m-{}^{I}\mathbf b_g+
    \omega_i^{\mathrm{roll}}{}^{I}\mathbf h_i\bigr).
    \label{eq:wheel_center_velocity}
\end{equation}

Under first-order discretization over a prediction interval $\Delta t$,
the wheel-position process-noise covariance is computed as
${}^{G}\mathbf R_I{}^{I}\boldsymbol\Sigma_i
    ({}^{G}\mathbf R_I)^\top\Delta t$.
\textcolor{black}{Conventional contact-aided filters use predefined contact-velocity
    noise covariances~\cite{hartley2020contact}, which cannot adapt to terrain
    changes or slip that varies across wheels and directions.
    Therefore, we use CovNet to predict the ${}^{I}\boldsymbol\Sigma_i$
    from proprioceptive history to estimate the uncertainty of each wheel's
    rolling motion in different directions (Sec.~\ref{sec:learned_uncertainty}),
    allowing the filter to adapt to changing wheel--terrain interactions.}

\subsubsection{Measurement Update}
Forward kinematics maps $\mathbf q^{\mathrm{leg}}$ to the wheel-center
position relative to the IMU origin,
${}^{I}\mathbf z_i=\mathbf f_i(\mathbf q^{\mathrm{leg}})$.
Its measurement model is
\begin{equation}
    {}^{I}\mathbf z_i
    ={}^{G}\mathbf R_I^{\top}
    ({}^{G}\mathbf p_{c_i}-{}^{G}\mathbf p_I)
    +{}^{I}\mathbf n_i^{\mathrm{FK}}.
    \label{eq:fk_measurement}
\end{equation}
Here, ${}^{I}\mathbf n_i^{\mathrm{FK}}\in\mathbb R^3$ denotes the
zero-mean FK position measurement noise for wheel $i$, expressed in the
IMU frame. Stacking these noise vectors for all four wheels gives a
12-dimensional noise vector with fixed covariance
$\mathbf C_{\mathrm{FK}}\in\mathbb R^{12\times12}$.
We apply the standard EKF update using this measurement model.
It links the wheel-center states to the IMU state and also corrects
velocity and biases through cross-covariances.
\textcolor{black}{The updated estimate posterior then serves as the prior
    for the LiDAR update in Sec.~\ref{sec:geometry_lidar}.}
\par

\subsection{Learning-Based Covariance}
\label{sec:learned_uncertainty}
\label{sec:differentiable_training}

CovNet maps proprioceptive history to the per-wheel process-noise
spectral densities used in Sec.~\ref{sec:wheel_process}.
For each wheel, CovNet predicts six scalar values that populate
a $3\times3$ lower-triangular matrix $\mathbf S_i$, yielding
\begin{equation}
    {}^{I}\boldsymbol\Sigma_i
    =\mathbf S_i\mathbf S_i^\top.
    \label{eq:learned_covariance}
\end{equation}
This parameterization ensures that each covariance matrix is
positive semidefinite.
The predicted covariances then serve as process-noise spectral
densities for wheel-center propagation.

The input to CovNet is a history of $H$ proprioceptive observations,
\begin{equation}
    \mathbf o:=\big(
    {}^{I}\boldsymbol\omega_m,{}^{I}\mathbf a_m,
    \mathbf q^{\mathrm{leg}},\dot{\mathbf q},\boldsymbol\tau,
    ({}^{I}\mathbf z_i,{}^{I}\dot{\mathbf z}_i)_{i=1}^{4}
    \big)\in\mathbb R^{74}.
    \label{eq:network_input}
\end{equation}
Here, $\boldsymbol\tau\in\mathbb R^{16}$ contains the joint efforts.
The FK positions ${}^{I}\mathbf z_i$ are defined in
Sec.~\ref{sec:wheel_process}, and ${}^{I}\dot{\mathbf z}_i$ are their
time derivatives in the IMU frame, computed from leg kinematics and
measured joint rates. The input stacks these quantities for all four wheels.
We exclude internal filter state estimates from CovNet inputs
to separate covariance prediction from the filter dynamics and
avoid feedback loops between them.
We denote the history of $H$ inputs by $\mathbf O$. All the inputs are normalized
along the time dimension~\cite{baumgartner2026coco}.
The normalized history is flattened and processed by a multilayer
perceptron (MLP).

We train CovNet using truncated backpropagation through a
differentiable filter~\cite{baumgartner2026coco}.
Training data consist of proprioceptive measurements and reference
states from simulated environments.
During each rollout, CovNet predicts per-wheel process-noise
spectral densities, and the filter performs propagation and FK updates
as described in Sec.~\ref{sec:wheel_process}.
We compute state-estimation errors and backpropagate them through
the filter to update CovNet.

Specifically, we define the per-sample loss as the squared IMU-velocity
error expressed in the IMU frame:
{\color{black}
\begin{equation}
    \mathcal L_v
    =\left\|
    ({}^{G}\hat{\mathbf R}_I)^\top{}^{G}\hat{\mathbf v}_I
    -({}^{G}\mathbf R_I^{\mathrm{GT}})^\top{}^{G}\mathbf v_I^{\mathrm{GT}}
    \right\|_2^2.
    \label{eq:velocity_loss}
\end{equation}
We minimize this loss averaged over training samples.
Here, ${}^{G}\hat{\mathbf R}_I$ and ${}^{G}\hat{\mathbf v}_I$ are the
IMU attitude and velocity estimated by the EKF, while the superscript
$\mathrm{GT}$ denotes the corresponding ground-truth values from the simulator.
We use body-frame velocity error because it is independent of the global
position and heading reference~\cite{baumgartner2026coco}.
Backpropagating this error through the EKF trains CovNet to predict
wheel-process uncertainties that improve velocity estimation, without
requiring covariance targets.
\par}

After training, CovNet is frozen and deployed on the real robot
without fine-tuning, using the same proprioceptive input definitions.
The full estimator additionally incorporates LiDAR updates,
with the FK posterior serving as the prior for each subsequent
LiDAR update.
Network configuration and training and deployment settings
are given in Sec.~\ref{sec:exp_parameters}.
\par

\subsection{LiDAR Update}
\label{sec:geometry_lidar}
\label{sec:degeneracy_update}

For LiDAR updates, we follow FAST-LIO2~\cite{xu2022fastlio2} for deskewing, point-to-plane registration, and the iterated Kalman update.
For each LiDAR point ${}^{I}\mathbf y$, let ${}^{G}\mathbf n_L$ and
${}^{G}\mathbf m$ denote the unit normal and a reference point of its
matched map plane, respectively. The point-to-plane residual is:
\begin{equation}
    \zeta(\mathbf x)=({}^{G}\mathbf n_L)^\top
    ({}^{G}\mathbf R_{I}{}^{I}\mathbf y+{}^{G}\mathbf p_{I}-{}^{G}\mathbf m).
    \label{eq:lidar_residual}
\end{equation}

Since LiDAR corrections can be unreliable in geometrically
weakly constrained directions, we introduce degeneracy-aware projection
(DAP) to suppress direct LiDAR correction in these directions.
\textcolor{black}{At each iteration, we recompute point-to-plane
    correspondences and form the stacked residual vector
    $\boldsymbol\zeta_L\in\mathbb R^{N_L}$ and its Jacobian
    $\mathbf H_{\mathrm{pose}}$ with respect to the six pose-error components
    of $\delta\mathbf x$. Here, $N_L$ is the number of matched point-to-plane
    pairs.} \textcolor{black}{With a fixed measurement noise variance
    $\sigma_L^2$ for all point-to-plane residuals, the Gauss--Newton information
    matrix admits the eigendecomposition (EVD)}
\begin{equation}
    \mathbf A_L
    ={\frac{1}{\sigma_L^2}
    \mathbf H_{\mathrm{pose}}^\top\mathbf H_{\mathrm{pose}}}
    =\mathbf V\boldsymbol\Lambda\mathbf V^\top,
    \label{eq:lidar_information}
\end{equation}
where $\boldsymbol\Lambda=\operatorname{diag}(\lambda_1,\ldots,\lambda_6)$,
$\lambda_1\leq\cdots\leq\lambda_6$, and
$\mathbf V=[\mathbf d_1,\ldots,\mathbf d_6]$ is orthogonal.
\textcolor{black}{Small eigenvalues indicate weak geometric constraints along
    the corresponding pose directions, even when the information matrix is full
    rank~\cite{hu2026dcreg}. We retain directions with sufficiently
    large eigenvalues to suppress weakly constrained LiDAR information.}

\begin{figure}[!t]
    \centering
    \includegraphics[width=\columnwidth]{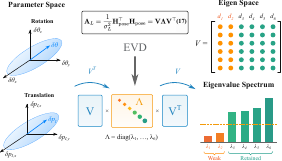}
    \caption{Joint six-dimensional analysis of LiDAR degeneracy through
        $\mathbf A_L=\mathbf V\boldsymbol\Lambda\mathbf V^\top$.
        Eigenvalue screening separates weak (orange) and retained (green)
        directions for Jacobian projection in DAP.
        \textcolor{black}{Here, $\theta$ denotes the local rotation
            coordinates of the IMU attitude, and $p_I$ denotes the
            IMU position in the global frame; $\delta$ denotes their perturbations.}
        Rotation and translation are drawn
        separately for clarity; the spectrum is illustrative.}
    \label{fig:lidar_observability}
\end{figure}

We retain $\mathcal S_{\mathrm{ret}}=\{s:\lambda_s\geq\gamma_\lambda\max(\lambda_6,\epsilon_\lambda)\}
    \cup\{6\}$, where $s\in\{1,\ldots,6\}$,
$\gamma_\lambda$ is a relative eigenvalue threshold and
$\epsilon_\lambda$ is a numerical floor. The largest-eigenvalue direction
is always retained to avoid an empty projector. DAP forms the projected Jacobian as
\begin{gather}
    \boldsymbol\Pi_{\mathrm{DAP}}
    =\sum_{s\in\mathcal S_{\mathrm{ret}}}\mathbf d_s\mathbf d_s^\top,
    \label{eq:observable_projector} \\
    \widetilde{\mathbf H}_L
    =\begin{bmatrix}
        \mathbf H_{\mathrm{pose}}\boldsymbol\Pi_{\mathrm{DAP}} &
        \mathbf 0_{N_L\times24}
    \end{bmatrix},
    \label{eq:projected_lidar_jacobian}
\end{gather}
Residuals remain unchanged; DAP suppresses direct LiDAR information in weak
directions~\cite{zhang2016degeneracy} (Fig.~\ref{fig:lidar_observability}).

The iterated update combines $\widetilde{\mathbf H}_L$ with the joint
IMU--wheel/FK prior. Learned wheel uncertainty controls the rolling contribution
to this prior, while DAP limits weak geometric information.
\textcolor{black}{This fusion allows reliable wheel motion}
to support weak geometry and LiDAR to limit drift when rolling is unreliable.
\par

\section{Experiments}
\label{sec:experiments}

\textcolor{black}{In this section, we conduct extensive experiments to evaluate
  the accuracy, robustness, and computational efficiency of the proposed
  estimator. Specifically, we examine how learned wheel uncertainty supports
  odometry under weak LiDAR geometry and how LiDAR limits drift when rolling
  becomes unreliable. Outdoor mapping further illustrates the estimator's
  performance across complex and varied real-world terrain.}

\subsection{Experimental Setup}
\label{sec:exp_setup}
\label{sec:exp_parameters}

\subsubsection{Platform, Data, and Training}
\begin{figure}[!t]
  \centering
  \includegraphics[width=\columnwidth,trim=0 0 0 4bp,clip]{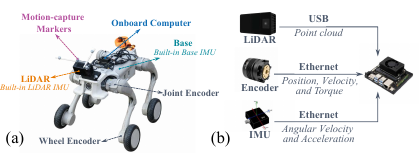}
  \caption{\textcolor{black}{Experimental setup.
    (a) Experimental platform: a quadrupedal wheel-legged robot equipped with
    an Odin1 module (Manifold Tech) and a Jetson Orin NX onboard computer.
    We use the module's integrated LiDAR, with a depth resolution of
    $240\times180$ pixels and a field of view of $120^{\circ}\times90^{\circ}$.
    (b) Communication among sensors: Raw encoder and IMU measurements are
    synchronized by the robot's internal low-level controller into a
    100~Hz data stream, while raw LiDAR data are provided at 10~Hz.}}
  \label{fig:exp_platform}
\end{figure}

Our platform (Fig.~\ref{fig:exp_platform}) carries an Odin1 LiDAR
with an integrated IMU, a base IMU, joint/wheel encoders, and a Jetson
Orin NX. The Odin1 IMU serves inertial propagation and deskewing;
the base IMU supplies CovNet inputs. LiDAR runs at 10~Hz.
\textcolor{black}{A low-level controller housed in the robot base synchronizes
  IMU and encoder measurements in software to form a 100~Hz proprioceptive
  data stream.}
The synchronized proprioceptive stream is sampled at 50~Hz for CovNet
inference.
\textcolor{black}{Each input is normalized over the
  history window using its mean and population standard deviation, with a
  standard-deviation floor of $10^{-5}$ to avoid division by zero.
  The network outputs are used directly to form $\mathbf S_i$ for each wheel.}

\textcolor{black}{Given the lack of suitable public datasets for quadrupedal
  wheel-legged robots, we collect our own dataset for real-world evaluation.
  It contains 19 indoor sequences (10 on wooden flooring and 9 on foam mats;
  50--120~s each), one corridor sequence with severe geometric degeneracy, four
  slip/disturbance sequences (approximately 300~s total), and a five-minute
  outdoor trajectory named \emph{wild}. Motion capture (Mocap) at 200~Hz
  provides indoor ground truth poses for evaluation. We plan to make this dataset publicly
  available.}

\textcolor{black}{CovNet is trained offline on a dataset collected in advance
  in Isaac Lab, comprising proprioceptive measurements and ground-truth robot
  states provided by the simulator. The dataset contains 16,384 sequences of
  192 time steps each. The sequences cover many randomized terrains with
  varying friction coefficients and surface roughness. CovNet is trained with
  \eqref{eq:velocity_loss} for 5,000 iterations on a single NVIDIA RTX 5090
  GPU, taking approximately 5 hours.
  Table~\ref{tab:estimator_settings} summarizes the training and deployment
  settings.}

\begin{figure*}[!tb]
  \centering
  \includegraphics[width=\textwidth]{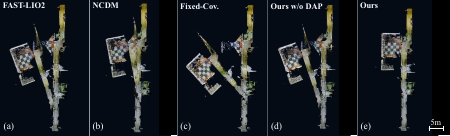}
  \caption{Corridor maps: (a) FAST-LIO2; (b) NCDM;
    (c) Fixed-Cov.; (d) Ours without DAP; (e) Ours.
    We use the
    internally colorized point cloud produced by
    Odin1 for visualization only. All evaluated methods use the raw
    point cloud output of the Odin1 LiDAR as input.
    \textcolor{black}{Only Ours avoids pronounced global drift
    of the corridor map, yielding the most consistent reconstruction
    among the compared methods.}}
  \label{fig:exp_corridor}
\end{figure*}

\begin{figure}[!t]
  \centering
  \includegraphics[width=\columnwidth]{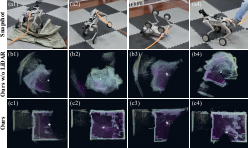}
  \caption{Rolling-model failures: (a1) slip on a smooth plastic sack;
    (a2) tether dragging; (a3) a kick lifting all wheels; (a4) a kick causing
    sliding. Corresponding maps use (b1)--(b4) Ours without LiDAR and
    (c1)--(c4) Ours. LiDAR points in (b) serve only to visualize estimated poses.}
  \label{fig:exp_disturbance}
\end{figure}

\begin{figure*}[!t]
  \centering
  \includegraphics[width=0.9\textwidth]{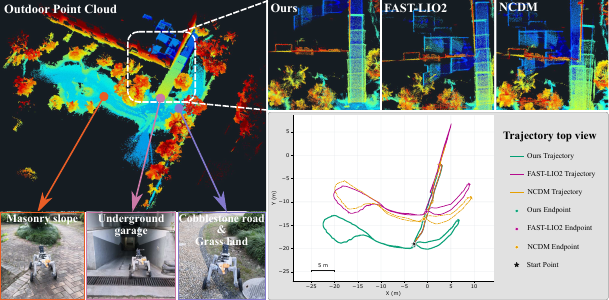}
  \caption{Outdoor mapping on mixed terrain. The large map uses Ours;
    upper-right panels compare Ours, FAST-LIO2, and NCDM.
    Photographs show terrain along the route; the lower-right panel compares
    their top-down trajectories.
    The route approximately returns to
    its starting location; because Mocap is unavailable outdoors, the point
    clouds and top-down trajectories provide a qualitative comparison.
    \textcolor{black}{Our method produces the most geometrically consistent map,
    particularly in the underground garage shown in the upper-right close-ups:
    FAST-LIO2 and NCDM exhibit visibly misaligned, duplicated walls,
    whereas our method preserves a coherent wall structure.}}
  \label{fig:exp_outdoor}
\end{figure*}

\begin{table}[htbp]
  \centering
  \caption{Training and deployment settings.}
  \label{tab:estimator_settings}
  \small
  \renewcommand{\arraystretch}{1.05}
  \setlength{\tabcolsep}{8pt}
  \begin{tabularx}{0.85\columnwidth}{@{}>{\raggedright\arraybackslash}X >{\centering\arraybackslash}p{0.22\columnwidth}@{}}
    \toprule
    \textbf{Setting}                           & \textbf{Value}                     \\
    \midrule
    \textcolor{black}{Input history length $H$} & \textcolor{black}{20}               \\
    \textcolor{black}{Hidden layer widths}      & \textcolor{black}{(128, 64)}        \\
    \textcolor{black}{Activation}               & \textcolor{black}{ELU}              \\
    \textcolor{black}{Parallel environments}    & \textcolor{black}{4,096}            \\
    \textcolor{black}{Batch size}               & \textcolor{black}{512}              \\
    \textcolor{black}{Learning rate}            & \textcolor{black}{$3\times10^{-5}$} \\
    \textcolor{black}{Optimizer}                & \textcolor{black}{Adam}             \\
    \textcolor{black}{Gradient-norm clipping}   & \textcolor{black}{1.0}              \\
    \bottomrule
  \end{tabularx}
\end{table}

\subsubsection{Baselines and Ablations}
\textcolor{black}{We compare our method with FAST-LIO2~\cite{xu2022fastlio2},
  a state-of-the-art LiDAR--inertial odometry method, and
  NCDM~\cite{zhou2025multimode}, a model-based odometry method that fuses LiDAR,
  IMU, and encoder measurements for quadrupedal wheel-legged robots.
  Since the original NCDM implementation is not publicly
  available, we reproduce it using reported joint efforts as torque proxies
  to estimate a contact probability for each wheel. A hysteresis rule
  identifies stance and swing, with increased process and observation
  covariances assigned during swing. All methods use the same recorded sensor
  data, with FAST-LIO2 using only LiDAR and IMU measurements. Mocap
  measurements are used for evaluation only.}

We also conduct two
ablations on the paper's core mechanisms. Fixed-Cov. replaces the learned
\textcolor{black}{IMU-frame wheel-process spectral density} with
\textcolor{black}{${}^{I}\boldsymbol\Sigma_i=0.003\mathbf I_3$}.
\textcolor{black}{Among the fixed values tested on the evaluation sequences,
  0.003 yielded the best overall performance.} Ours without DAP disables the
projection by setting $\boldsymbol\Pi_{\mathrm{DAP}}=\mathbf I_6$, thereby
retaining the full LiDAR Jacobian. The same value also supplies NCDM's base
covariance. All other settings are
retained.

\subsubsection{Evaluation Protocol}
\textcolor{black}{For the room and slip/disturbance sequences, we align each
  estimate to Mocap at the first common pose and report position ATE RMSE.
  The corridor starts and ends in the Mocap room; its end-to-end (E2E)
  position error uses Mocap poses recorded before departure and after return.
  The outdoor \emph{wild} run lacks Mocap ground truth and only approximately
  returns to its start, so we compare point-cloud maps and top-down
  trajectories qualitatively (Fig.~\ref{fig:exp_outdoor}) without reporting a
  quantitative E2E error.}

\subsection{Odometry under Geometric Degeneracy}
\label{sec:exp_accuracy}
\label{sec:exp_corridor}

\textcolor{black}{We evaluate odometry as degeneracy becomes more frequent,
from room sequences to the corridor (Table~\ref{tab:exp_mocap}). Following
DCReg~\cite{hu2026dcreg}, we define the degeneracy rate (DR) as
$\mathrm{DR}=100N_{\mathrm{deg}}/N_{\mathrm{valid}}$ (\%), where
$N_{\mathrm{valid}}$ counts frames with valid degeneracy assessments and
$N_{\mathrm{deg}}$ those with at least one direction flagged as degenerate.
Initialization frames and invalid assessments are excluded; the difficulty
groups reflect DR.}

\begin{table}[!t]
  \centering
  \caption{\textcolor{black}{Room ATE RMSE and corridor E2E position error (m).}}
  \label{tab:exp_mocap}
  \small
  \setlength{\tabcolsep}{2pt}
  \renewcommand{\arraystretch}{1.05}
  \begin{tabular*}{\columnwidth}{@{\extracolsep{\fill}}l r r r r r r@{}}
    \toprule

    Seq. & DR (\%) & \makecell{FAST-\\LIO2}
    & NCDM
    & \makecell{Fixed-\\Cov.}
    & \makecell{Without\\DAP} & Ours \\
    \midrule
    \multicolumn{7}{@{}l}{\textit{Easy ($\mathrm{DR}=0$)}} \\
    \addlinespace[2pt]
    00 & 0.000 & 0.079 & 0.082 & 0.079 & 0.075 & \textbf{0.075} \\
    01 & 0.000 & 0.189 & 0.197 & 0.187 & 0.153 & \textbf{0.153} \\
    02 & 0.000 & 0.167 & 0.175 & 0.179 & 0.167 & \textbf{0.167} \\
    03 & 0.000 & 0.087 & 0.093 & 0.090 & 0.071 & \textbf{0.071} \\
    04 & 0.000 & 0.064 & \textbf{0.037} & 0.040 & 0.040 & 0.040 \\
    05 & 0.000 & 0.042 & 0.044 & 0.042 & 0.030 & \textbf{0.030} \\
    \midrule
    \multicolumn{7}{@{}l}{\textit{Medium ($0<\mathrm{DR}\leq5\%$)}} \\
    \addlinespace[2pt]
    06 & 0.162 & 0.120 & 0.117 & 0.114 & \textbf{0.076} & 0.077 \\
    07 & 0.165 & 0.085 & 0.087 & 0.084 & 0.088 & \textbf{0.081} \\
    08 & 0.324 & 0.406 & 0.420 & 0.417 & 0.310 & \textbf{0.310} \\
    09 & 1.294 & \textbf{0.069} & 0.080 & 0.079 & 0.071 & 0.071 \\
    10 & 1.942 & 0.088 & 0.085 & 0.084 & 0.084 & \textbf{0.081} \\
    11 & 2.760 & 0.289 & 0.094 & 0.122 & 0.078 & \textbf{0.078} \\
    12 & 3.510 & \textbf{0.204} & 0.205 & 0.249 & 0.220 & 0.220 \\
    \midrule
    \multicolumn{7}{@{}l}{\textit{Hard ($5\%<\mathrm{DR}\leq50\%$)}} \\
    \addlinespace[2pt]
    13 & 5.502 & 0.133 & 0.132 & 0.130 & 0.118 & \textbf{0.112} \\
    14 & 7.455 & 0.089 & 0.084 & 0.082 & 0.079 & \textbf{0.076} \\
    15 & 7.605 & 0.366 & 0.511 & 0.346 & 0.184 & \textbf{0.183} \\
    16 & 8.414 & 0.541 & 0.169 & 0.167 & 0.099 & \textbf{0.099} \\
    17 & 13.636 & 0.448 & 0.420 & 0.388 & 0.302 & \textbf{0.301} \\
    18 & 20.131 & 1.056 & 0.178 & 0.179 & 0.120 & \textbf{0.120} \\
    \midrule
    \multicolumn{7}{@{}l}{\textit{Extremely hard ($\mathrm{DR}>50\%$); E2E error}} \\
    \addlinespace[2pt]
    corridor & 60.832 & 7.531 & 2.801 & 12.042 & 15.084 & \textbf{0.198} \\
    \bottomrule
  \end{tabular*}
  \par\smallskip
\end{table}

\textcolor{black}{The proposed method remains competitive on Easy and Medium room sequences
  and has lower RMSE than FAST-LIO2, NCDM, and Fixed-Cov. on all six Hard
  sequences. Its mean RMSE in the Hard group is 0.148~m, 66.2\% lower than
  FAST-LIO2 and 31.0\% lower than Fixed-Cov. Removing DAP barely changes these results, indicating that
  learned wheel uncertainty accounts for most of this gain.}

\textcolor{black}{The corridor makes DAP's contribution visible
  ($\mathrm{DR}=60.832\%$; Fig.~\ref{fig:exp_corridor}). Ours achieves
  0.198~m E2E error, versus 2.801~m for NCDM, 12.042~m for Fixed-Cov.,
  and 15.084~m for Ours without DAP. These ablations suggest complementary
  roles in this run: learned uncertainty regulates the rolling prior, while
  DAP limits corrections along weak LiDAR directions.}

\subsection{Robustness to Unreliable Wheel Motion}
\label{sec:exp_disturbance}

\textcolor{black}{After testing weak LiDAR geometry, we turn to sequences in
  which wheel motion is unreliable. Slip on a smooth plastic sack, tether
  dragging, a kick that lifts all wheels, and sliding after a kick challenge the
  rolling model in different ways (Fig.~\ref{fig:exp_disturbance}). We compare
  Ours with Ours without LiDAR, which retains the same IMU and wheel/FK
  estimation pipeline but removes LiDAR updates. Without LiDAR, position RMSE
  ranges from 0.640 to 1.439~m;
  with LiDAR fusion, it falls to 0.057--0.220~m, with lower error on all four
  sequences (Table~\ref{tab:exp_disturbance}). The largest reduction occurs on
  \emph{kick00}, from 1.439~m to 0.061~m, when a kick lifts all wheels. The
  corresponding maps show distorted room boundaries without LiDAR and clearer
  structure with fusion. These results show that LiDAR limits drift when rolling
  predictions fail, complementing the wheel/FK prior's role under weak LiDAR
  geometry in Sec.~\ref{sec:exp_accuracy}.}

\begin{table}[!htbp]
  \centering
  \caption{Position RMSE under slip and disturbances (m).}
  \label{tab:exp_disturbance}
  \small
  \setlength{\tabcolsep}{6pt}
  \begin{tabular}{@{}l r r r r@{}}
    \toprule
    Method             & \emph{slip00}  & \emph{slip01}  & \emph{kick00}  & \emph{kick01}  \\
    \midrule
    Ours without LiDAR & 0.844          & 0.640          & 1.439          & 1.014          \\
    Ours               & \textbf{0.125} & \textbf{0.220} & \textbf{0.061} & \textbf{0.057} \\
    \bottomrule
  \end{tabular}
\end{table}

\subsection{Outdoor Mapping and Computational Cost}
\label{sec:exp_outdoor}
\label{sec:exp_ablation_runtime}

\textcolor{black}{We next examine
  the estimator over a longer outdoor route with varied terrain and geometry.
  The five-minute \emph{wild} trajectory crosses masonry slopes, a narrow
  garage entrance, cobblestones, and grass before approximately returning to
  its starting area. Ours produces a visually coherent point-cloud map across
  the route (Fig.~\ref{fig:exp_outdoor}). The estimated top-down paths and
  endpoints differ across methods. Without outdoor Mocap ground truth, these
  maps and trajectories provide a qualitative comparison but do not establish an
  accuracy ranking.}

\begin{figure}[!tb]
  \centering
  \includegraphics[width=\columnwidth]{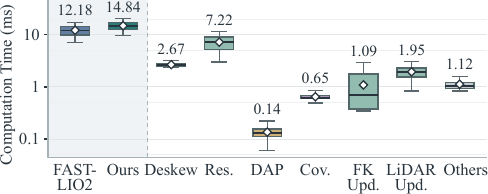}
  \caption{CPU time per scan: totals and breakdown of Ours. Deskew:
    IMU propagation/deskewing; Res.: point-to-plane residual construction;
    DAP: degeneracy-aware projection; Cov.: network inference; FK Upd.: FK
    reconstruction/wheel update; LiDAR Upd.: remaining iterative update;
    Others: remaining computation. Diamonds/numbers show means, lines
    medians, boxes interquartile ranges, and whiskers 5th--95th percentiles.}
  \label{fig:exp_runtime}
\end{figure}

\textcolor{black}{We profile computational cost on the same \emph{wild} run
  using a single logical CPU core on AMD Ryzen AI 9 HX 370 without GPU. Three replays per method yield 9,441 scan windows
  after discarding 50 warm-up frames per replay. Timings cover estimation,
  network inference, FK reconstruction, and amortized preparation; they
  exclude data loading, initialization, interprocess transfer, logging, and
  export. Under this setup, Ours averages 14.84~ms per scan, compared with
  12.18~ms for FAST-LIO2, an increase of 2.66~ms
  (Fig.~\ref{fig:exp_runtime}). Covariance inference and DAP account for
  0.649~ms and 0.136~ms, respectively, while point-to-plane residual
  construction remains the largest component at 7.22~ms. The measured mean
  is below the 100~ms interval of the 10~Hz LiDAR.}

\section{Conclusion}
\label{sec:conclusion}

We presented a tightly coupled LiDAR--inertial--kinematic odometry
framework for quadrupedal wheel-legged robots. Rather than relying on binary
contact states, the method learns per-wheel, direction-dependent process
covariances from proprioceptive history and uses them to regulate analytical
rolling constraints through FK corrections. CovNet is trained
through a differentiable filter using simulated velocity supervision, without
contact labels, and is deployed on the real robot without fine-tuning. DAP
complements the learned wheel uncertainty by suppressing direct LiDAR
information along geometrically weak directions. On the six room sequences
with the highest degeneracy ratios, the method reduces mean position RMSE by
66.2\% relative to FAST-LIO2 and by 31.0\% relative to the fixed-covariance
variant. In the severely degenerate corridor sequence, it achieves an
endpoint error of 0.198~m.

\textcolor{black}{The current experiments primarily evaluate geometric degeneracy
and unreliable wheel motion separately, leaving robustness under their severe
co-occurrence unverified. CovNet is trained exclusively in simulation, and
the benefits of real-world training or adaptation have not been evaluated.
Future work will explore how the estimated wheel-center
positions and learned wheel-motion uncertainty can support uncertainty-aware
foothold planning, linking the proposed state estimation framework to
motion planning for wheel-legged robots.}

\section*{Acknowledgment}
This work was supported in part by the InnoHK initiative of the Innovation
and Technology Commission of the Hong Kong Special Administrative Region
Government via the Hong Kong Centre for Logistics Robotics. This work was
also supported in part by project MMT-p8-26 of the Shun Hing Institute of
Advanced Engineering, The Chinese University of Hong Kong.

\bibliographystyle{IEEEtran}
\IEEEtriggeratref{26}
\bibliography{references}
\end{document}